\documentclass[11pt]{article}

\usepackage[preprint]{acl}

\usepackage{times}
\usepackage{latexsym}

\usepackage[T1]{fontenc}

\usepackage[utf8]{inputenc}

\usepackage{microtype}

\usepackage{inconsolata}

\usepackage{graphicx}

\usepackage{multirow}
\usepackage{amssymb}
\usepackage{makecell}
\usepackage{amsmath}
\usepackage{relsize}
\usepackage{arydshln}
\usepackage{booktabs}
\usepackage{bm}
\usepackage{hyperref}

\usepackage{url}

\usepackage{amsfonts}
\usepackage{graphicx}
\usepackage{algorithm}
\usepackage{algorithmic}
\usepackage{tabu}
\usepackage{subfigure}

\usepackage{xcolor}
\usepackage{wrapfig} 
\usepackage{colortbl}
\usepackage[most]{tcolorbox}

\usepackage[normalem]{ulem}
\useunder{\uline}{\ul}{}

\definecolor{wkyellow}{RGB}{255,241,177}
\definecolor{lightgray}{HTML}{CCE5FF}
\definecolor{lightgray}{gray}{0.9}
\definecolor{goodblue}{HTML}{0071bc}

\title{Revealing the Learning Dynamics of Long-Context Continual Pre-training}

\author{
 \textbf{Yupu Liang\textsuperscript{1}}\thanks{Equal contribution.},
 \textbf{Shuang Chen\textsuperscript{1$\ast$}},
 \textbf{Guanwei Zhang\textsuperscript{1}},
 \textbf{Shaolei Wang\textsuperscript{1}},
 \textbf{Suncong Zheng\textsuperscript{1}}\thanks{Corresponding Author.}
\\
 \textsuperscript{1}Tencent Hunyuan, Beijing, China
\\
 \small{
 \texttt{\{yupuling, alexschen, waingwzhang, robertswang, congzheng\}@tencent.com}
 }
 \makeatletter{\footnotetext{Equal contribution.}}\makeatother
}

\begin{document}
\maketitle
\begin{abstract}

Existing studies on Long-Context Continual Pre-training (LCCP)  mainly focus on small-scale models and limited data regimes (tens of billions of tokens). We argue that directly migrating these small-scale settings to industrial-grade models risks insufficient adaptation and premature training termination. Furthermore, current evaluation methods rely heavily on downstream benchmarks (e.g., Needle-in-a-Haystack), which often fail to reflect the intrinsic convergence state and can lead to "deceptive saturation".
In this paper, we present \textit{the first systematic investigation of LCCP learning dynamics using the industrial-grade Hunyuan-A13B} (80B total parameters), tracking its evolution across a 200B-token training trajectory.
Specifically, we propose a hierarchical framework to analyze LCCP dynamics across \textbf{behavioral} (supervised fine-tuning probing), \textbf{probabilistic} (perplexity), and \textbf{mechanistic} (attention patterns) levels.
Our findings reveal:
(1) \textbf{Necessity of Massive Data Scaling:} Training regimes of dozens of billions of tokens are insufficient for industrial-grade LLMs' LCCP (e.g., Hunyuan-A13B reaches saturation after training over 150B tokens).
(2) \textbf{Deceptive Saturation vs. Intrinsic Saturation:} Traditional NIAH scores report "fake saturation" early, while our PPL-based analysis reveals continuous intrinsic improvements and correlates more strongly with downstream performance.
(3) \textbf{Mechanistic Monitoring for Training Stability:} Retrieval heads act as efficient, low-resource training monitors, as their evolving attention scores reliably track LCCP progress and exhibit high correlation with SFT results.
This work provides a comprehensive monitoring framework, evaluation system, and mechanistic interpretation for the LCCP of industrial-grade LLM.

\end{abstract}

\section{Introduction}

The evolution of Large Language Models (LLMs) has recently shifted from scaling parameter counts to extending effective context windows, enabling applications such as long-document comprehension \citep{ding2024longrope}, complex code synthesis \citep{le2025impacts}, and sophisticated retrieval-augmented generation \citep{jiang2024longrag}. Endowing models with the ability to process long-range dependencies typically involves a specialized phase known as \textbf{Long-Context Continual Pre-training (LCCP)}. While architectural innovations such as Rotary Positional Embedding (RoPE) interpolation \citep{su2024roformer} and Grouped-Query Attention (GQA) \citep{ainslie2023gqa} provide the structural foundation for sequence extension, the learning dynamics of LCCP particularly within industrial production environments, remain under-explored.

Existing studies on LCCP primarily focus on small-scale academic models (e.g., 7B parameters) trained on limited data regimes (e.g., 20B tokens), typically ranging from a few billion to tens of billions of tokens \citep{gao2025train, retrieval2025, zhong2025understanding}. We argue that this creates a significant discrepancy between academic findings and industrial reality. First, conclusions drawn from small-scale models may fail to generalize to industrial-grade models. For instance, training data volumes that reach saturation in smaller settings may remain undersaturated as model capacity increases. Second, current evaluation methodologies rely heavily on downstream benchmarks like the "Needle-In-A-Haystack" (NIAH) test \citep{niah2023}. While NIAH provides a coarse-grained measure of retrieval, it often reports "deceptive saturation", a phenomenon where the metric suggests the model has converged, yet fails to capture the fine-grained refinements occurring in long-context modeling.

In this paper, we present \textit{the first systematic investigation of LCCP learning dynamics in an industrial-grade production model}: Hunyuan-A13B, a Sparse Mixture-of-Experts (MoE) model with 80B total parameters \citep{hunyuan_a13b_report}. We track the model's evolution across a 200B-token LCCP trajectory, extending its context capacity from 32K to 64K. 
The cumulative computational cost of LCCP stage is estimated at $4 \times 10^{23}$ total floating-point operations (FLOPs).

Besides, we also propose a hierarchical analysis framework that investigates LCCP through three distinct levels:
(1) \textbf{Behavioral Level:} Utilizing lightweight Supervised Fine-Tuning (SFT) probing to evaluate the model's long-context understanding abilities in downstream tasks.
(2) \textbf{Probabilistic Level:} Analyzing fine-grained Perplexity (PPL) and continuous NIAH score to capture the intrinsic convergence state.
(3) \textbf{Mechanistic Level:} Identifying the evolution of internal attention patterns and the emergence of specialized "retrieval heads".

Our investigation yields three critical insights into the nature of long-context adaptation:
\begin{itemize}
    \item \textbf{Necessity of Massive Data Scaling:} We demonstrate that for industrial-grade LLMs, conventional data scales (dozens of billions of tokens) are insufficient. Our results show that Hunyuan-A13B requires over 150B tokens to reach a stable saturation plateau, suggesting that LCCP is more data-intensive than previously estimated.
    \item \textbf{Identifying Deceptive Saturation:} We reveal that NIAH scores often provide a "fake saturation" signal. While NIAH may plateau early, our PPL-based analysis uncovers continuous, intrinsic improvements in generation confidence that correlate more strongly with actual downstream performance.
    \item \textbf{Retrieval Heads as Efficient Monitors:} We establish a strong positive correlation between the retrieval scores of specific "retrieval heads" and model proficiency. This mechanistic discovery provides a low-resource, rapid diagnostic tool for monitoring LCCP progress without the need for exhaustive benchmarking.
\end{itemize}

This work provides a comprehensive monitoring framework, evaluation system, and mechanistic interpretation for the LCCP of industrial-grade models. By bridging the gap between academic theory and industrial-grade implementation, we offer a roadmap for more efficient and predictable long-context adaptation and evaluation in LLMs.

\section{Related Work}

\noindent \textbf{Pre-trained Model Evaluation:} Pre-training defines the performance ceiling of LLMs \citep{yue2025does}. However, evaluating base models is challenging due to their limited instruction-following capabilities and the high computational costs of training. To address this, existing frameworks like lm-evaluation-harness \citep{eval-harness} and OpenCompass \citep{2023opencompass} predominantly rely on ICL. By providing few-shot examples, these methods steer models to generate structured responses without explicit instruction tuning \citep{dong2024survey, luan2025bose}.

\noindent \textbf{Long Context Evaluation:} Benchmarks for long-context modeling generally fall into two categories:
(1) Real-World Tasks: These assess practical utility using authentic data from domains such as document understanding \citep{karpinska2024one}, safety \citep{huang2024longsafetybench}, and medicine \citep{hosseini2024benchmark}.
(2) Synthetic Tasks: Benchmarks like RULER \citep{hsiehruler} and LongBench \citep{bai2025longbench} employ synthetic NIAH tests. These allow for fine-grained control over context length and complexity, facilitating a more intrinsic analysis of a model's retrieval and reasoning mechanisms \citep{das2024needle, vodrahalli2024michelangelo}.

In this work, we not only employ a variety of conventional long-context benchmarks to illustrate the learning dynamics of the pre-trained model, but also utilize samples from LongBioBench \citep{yang2025controllable}, combined with the metric of PPL, to conduct a more fine-grained, intrinsic analysis of the model's performance.

\section{LCCP Settings}

\subsection{Model Structure}

In this work, we adopt the Hunyuan-A13B architecture, an open-source Large Language Model engineered to balance computational efficiency with high-level reasoning capabilities \citep{hunyuan_a13b_report}. The detailed structural configurations are summarized in Table \ref{tab:model_config}. Hunyuan-A13B utilizes a Sparse Mixture-of-Experts design, comprising 80B total parameters with 13B activated per token. Each MoE layer integrates one shared expert and 64 specialized experts, the latter of which are dynamically routed to select 8 experts during inference. To enhance long-context processing, the architecture employs Grouped-Query Attention with 32 query heads and 8 KV heads, effectively minimizing KV cache overhead. Additionally, the model implements the SwiGLU activation function, with a hidden dimension of 4096 and an internal FFN hidden size of 3072 for each expert.

\begin{table}[t]
\centering
\caption{Key hyper-parameter of Hunyuan-A13B.}
\label{tab:model_config}
\small
\begin{tabular}{ll}
\toprule
\textbf{Hyper-parameter} & \textbf{Value} \\ \midrule
Total Parameters         & 80B            \\
Active Parameters        & 13B            \\
Layers                   & 32             \\
Hidden Size              & 4096           \\
FFN Hidden Size          & 3072           \\
Heads (Q / KV)           & 32 / 8         \\
Activation Function      & SwiGLU         \\
Shared Experts           & 1              \\
Specialized Experts      & 64             \\
Activated Specialized Experts & 8         \\ \bottomrule
\end{tabular}
\end{table}

\subsection{Data Mix}
The quality and diversity of the training corpus are pivotal for successful LCCP. In this work, we employ a hybrid data strategy designed to balance the acquisition of long-dependency modeling capabilities with the preservation of general language proficiency. Our data mixture consists of 25\% short-context data (following the same distribution as the initial pre-training stage) and 75\% long-context documents with lengths exceeding 32k tokens.

The composition of the long-context portion is meticulously curated from multiple high-quality sources, as summarized in Table \ref{tab:data_distribution}. Specifically, Common Crawl (36.3\%) and Books (28.6\%) provide broad linguistic knowledge and diverse narrative structures, while arXiv (24.0\%) and Code (10.8\%) contribute dense, structured information essential for complex reasoning and long-range retrieval. This high proportion of technical and structured data ensures the model maintains robust performance during the context extension process.

\begin{table}[t]
\centering
\caption{Distribution of data sources for long-context documents.}
\label{tab:data_distribution}
\small
\begin{tabular}{lr}
\toprule
\textbf{Source}            & \textbf{Ratio (\%)} \\ \midrule
Common Crawl (CC)          & 36.3                \\
Books                      & 28.6                \\
arXiv                      & 24.0                \\
Code                       & 10.8                \\
Wikipedia                  & 0.3                 \\ \midrule
\textbf{Total}             & \textbf{100.0}      \\ \bottomrule
\end{tabular}
\end{table}

\subsection{Training Recipe}

We specifically utilize the Hunyuan-A13B-32k base model as our starting point, extending its context capacity from 32k to 64k through our LCCP stage on a total of 200 billion tokens.
To accommodate the expanded context, we adjust the RoPE base frequency, increasing it from 500K to 2M.

The model is trained using a constant learning rate of $1.2 \times 10^{-5}$ to ensure stable weight updates during the adaptation phase. We employ a global batch size of 16M tokens. This configuration allows the model to progressively adapt to extended sequences while mitigating the risk of catastrophic forgetting of the knowledge acquired during the initial pre-training phase.

\section{Behavioral Level Analysis}

In this section, we utilize the SFT probe to assess the long-context capabilities of our pre-trained models.
Following \citet{gao2025train}, we perform lightweight SFT on intermediate checkpoints during the LCCP phase, and subsequently evaluate these SFT-adapted models on standard long-context benchmarks to analyze the advancements of their long-context understanding abilities.

\begin{figure*}[t]
    \centering
    \includegraphics[width=2\columnwidth]{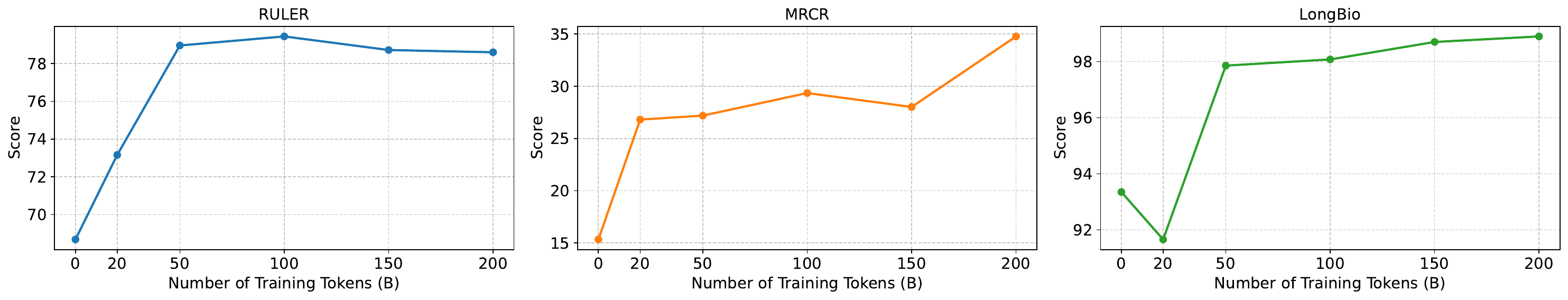}
    \caption{The performance on the RULER, MRCR, and LongBio benchmarks of different LCCP checkpoints of the Hunyuan-A13B model after lightweight SFT.}
    \label{figure: sft_results}
     
\end{figure*}

\subsection{SFT Training Recipe}

The SFT probes are conducted using a high-quality internal dataset comprising 440K samples, totaling approximately 0.25B tokens. To ensure a comprehensive evaluation, the dataset covers a broad spectrum of domains and tasks, including general capabilities, reasoning skills, and long-context specific tasks.

To maintain the "probing" nature of this evaluation and prevent over-fitting, we adopt a conservative training configuration. We employ a learning rate that decays from $2 \times 10^{-5}$ to $5 \times 10^{-6}$ using a cosine schedule. The model is fine-tuned for 2 epochs with a global batch size of 1M tokens.

\subsection{Performance Evolution}

After lightweight SFT, we conduct a comprehensive evaluation of various pre-trained checkpoints utilizing several prominent open-source long-context benchmarks.
Specifically, we evaluate our models on the RULER \citep{hsieh2024ruler}, MRCR \citep{Vodrahalli2024MichelangeloLC}, and LongBio \citep{Yang2025ACE} benchmarks.
For the RULER benchmark, we select its question answer subset because of the relevance to real-world scenarios.
For the LongBio benchmark, we select its paraphrase, pronoun, and standard subsets to assess models' understanding ability.
To ensure consistency, the sequence lengths of all evaluation samples are strictly maintained within the 32K to 64K range, characterized by a uniform distribution across this interval.
We report the Pass@3 scores for all checkpoints following SFT.
The results are shown in Figure~\ref{figure: sft_results}.

As illustrated in the figure, long-context performance scales with training volume, eventually saturating after 100B tokens, as evidenced by RULER scores rising from 68.68 to a peak of 79.44 at the 100B-token mark. While benchmarks like MRCR and LongBio exhibit diminishing marginal returns after 50B tokens, the overall trajectory reveals a critical insight: \textbf{contrary to academic studies that observe saturation at much smaller scales, industrial-grade LLMs require a significantly larger data regime, exceeding 100B tokens, to achieve stable and comprehensive context adaptation}.

\section{Probabilistic Level Analysis}

In this section, we modify the original NIAH task by transforming it from a result-based binary evaluation metric into a PPL-based continuous evaluation metric.
This PPL-based approach addresses the deceptive saturation in NIAH, providing a more fine-grained reflection of incremental improvements in the model's long-context performance.
Specifically, we reformulate the test texts of the original NIAH task, and compute the token-level PPL over the answer tokens for each sample.

\subsection{Continuous NIAH}
\label{section: niah}

\begin{figure*}[t]
    \centering
    \includegraphics[width=2\columnwidth]{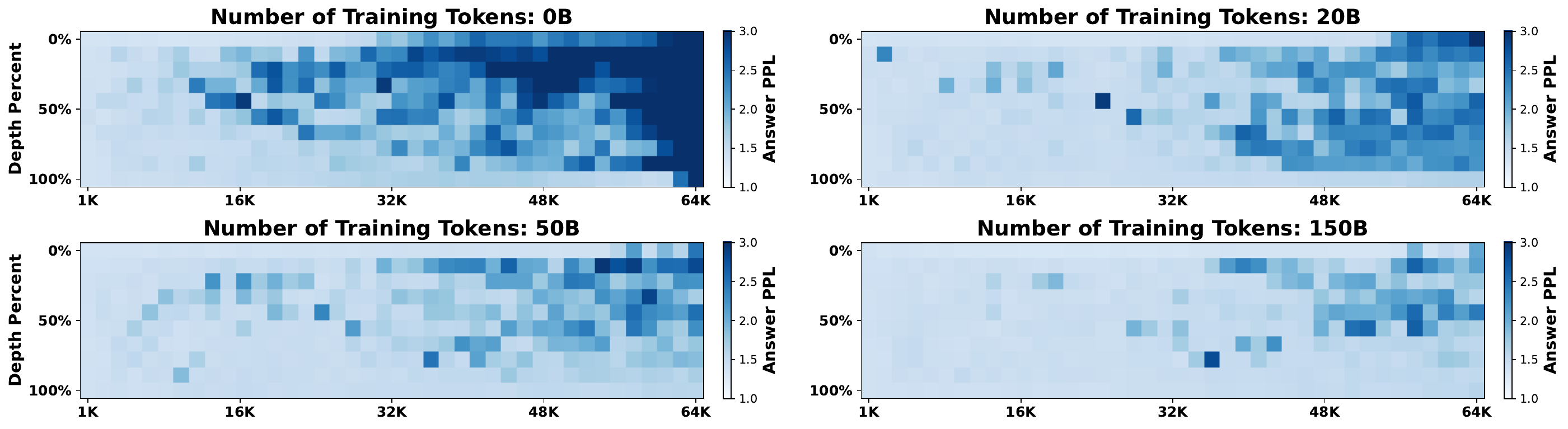}
    \caption{The PPL distribution for NIAH tasks across different context lengths and depths for Hunyuan-A13B base model at different stages of LCCP. Each cell represents the average PPL of the answer tokens.}
    \label{figure: ppl_map}
\end{figure*}

The original Needle-in-a-Haystack \citep{niah2023} is designed to quantitatively evaluate an LLM's ability to locate and recall specific information when processing long-sequence inputs.
However, this approach has two primary limitations: (1) For base models that have undergone LCCP without subsequent SFT, their suboptimal instruction-following performance may not serve as a faithful proxy for their long-context capabilities. (2) Traditional result-based binary evaluation metrics focus exclusively on the accuracy of generated answers, thereby overlooking the underlying token-level probability distribution; even when models produce identical outputs, their generation probabilities may diverge significantly.

Therefore, a more appropriate evaluation strategy is to feed the model a completion sequence that includes the context, the question, and the answer, and then compute the PPL over the answer tokens.
By converting the discrete NIAH scoring scheme into continuous PPL values, our approach enables the detection of subtle, incremental improvements in long-context capabilities that occur throughout the training process.

Specifically, we feed text in the following format into Hunyuan-A13B base model at different stages of LCCP.

\begin{tcolorbox}[colback=lightgray!50!white,colframe=lightgray,title=\textcolor{black}{Needle-in-a-Haystack Inputs}, width=\columnwidth, breakable]
\footnotesize
\texttt{[Context Before Needle Text]}
\\\\
The best thing to do in San Francisco is eat a sandwich and sit in Dolores Park on a sunny day.
\\\\
\texttt{[Context After Needle Text]}
\\\\
Question: What is the best thing to do in San Francisco?
\\\\
Answer: The best thing to do in San Francisco is eat a sandwich and sit in Dolores Park on a sunny day.
\end{tcolorbox}

By controlling the context length before and after the needle text, we evaluate total sequence lengths ranging from 1K to 64K and depths from 0\% to 100\%, and compute the average PPL of the tokens following \textbf{\texttt{Answer:}}

\subsection{Deceptive Saturation in NIAH}

\begin{figure}[t]
    \centering
    \includegraphics[width=0.9\columnwidth]{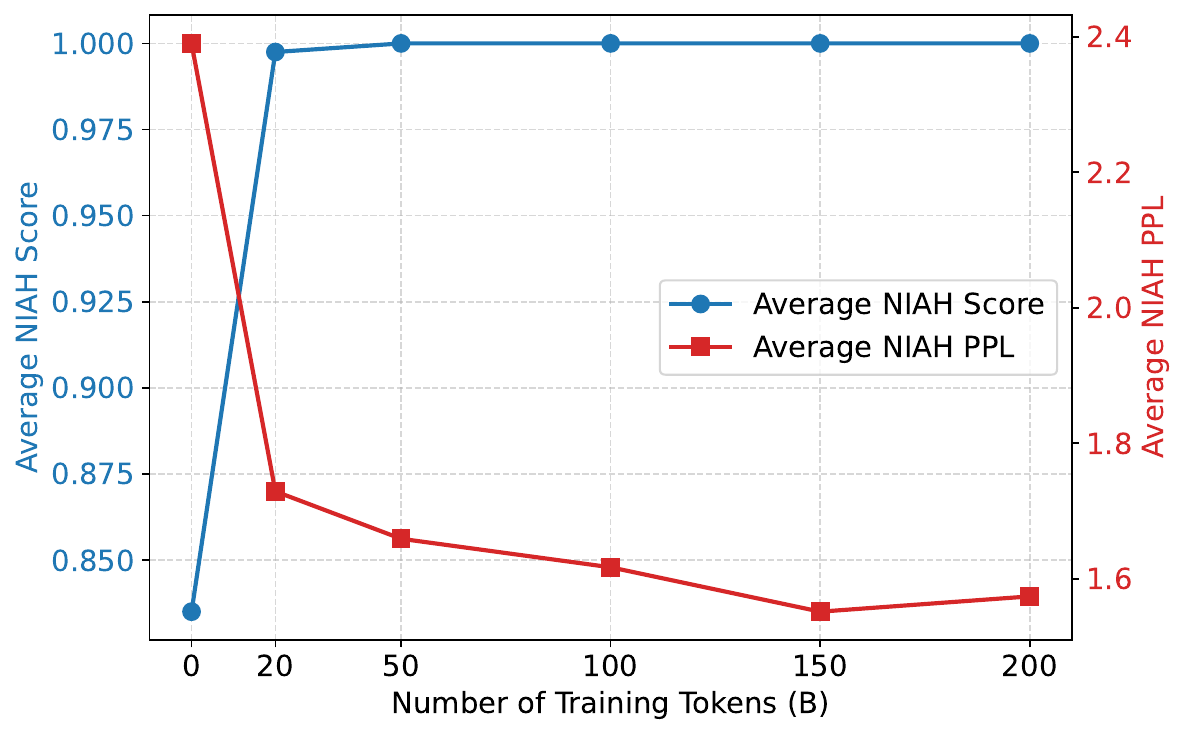}
    \caption{Average NIAH score and PPL across Hunyuan-A13B base model at different stages of LCCP.}
    \label{figure: combined_niah_score_ppl_curve}
\end{figure}

Figure~\ref{figure: ppl_map} presents the PPL heatmaps, providing a granular view of performance across varying context lengths and retrieval depths.
In addition, Figure~\ref{figure: combined_niah_score_ppl_curve} illustrates the average NIAH score and PPL for the Hunyuan-A13B base model across various LCCP stages.

In Figure~\ref{figure: combined_niah_score_ppl_curve}, the original NIAH score approaches saturation as early as 20B tokens and maintains a consistent 100\% accuracy beyond 50B tokens. In contrast, the PPL curve exhibits a sustained decline, only reaching a plateau at 150B tokens. Furthermore, correlation analysis with downstream SFT probe benchmark results (Table~\ref{tab:naih_ppl_correlation}) reveals that NIAH PPL yields a significantly higher correlation than the NIAH score.\footnote{The significance testing can be seen in Appendix~\ref{appendix: p-value} Table~\ref{tab:naih_ppl_correlation_p_value}.}
These findings provide empirical evidence that \textbf{the original NIAH metric suffers from deceptive saturation, whereas our proposed continuous NIAH serves as a more faithful indicator of the model's long-context modeling performance}.

\begin{table}[t]
\centering
\caption{The Pearson correlation coefficients between NIAH metrics and performance on downstream SFT probe benchmarks.}
\label{tab:naih_ppl_correlation}
\small
\begin{tabular}{lcc}
\toprule
\textbf{Metric} & \textbf{NIAH Score} & \textbf{NIAH PPL} \\ \midrule
RULER              & 0.8552              & -0.9115           \\
MRCR               & 0.8927              & -0.9231           \\
LongBio            & 0.4977              & -0.6283           \\ \midrule
\textbf{Average}   & \textbf{0.7486}     & \textbf{-0.8210}  \\ \bottomrule
\end{tabular}
\end{table}

\section{Mechanistic Level Analysis}

\begin{figure*}[t]
    \centering
    \includegraphics[width=2\columnwidth]{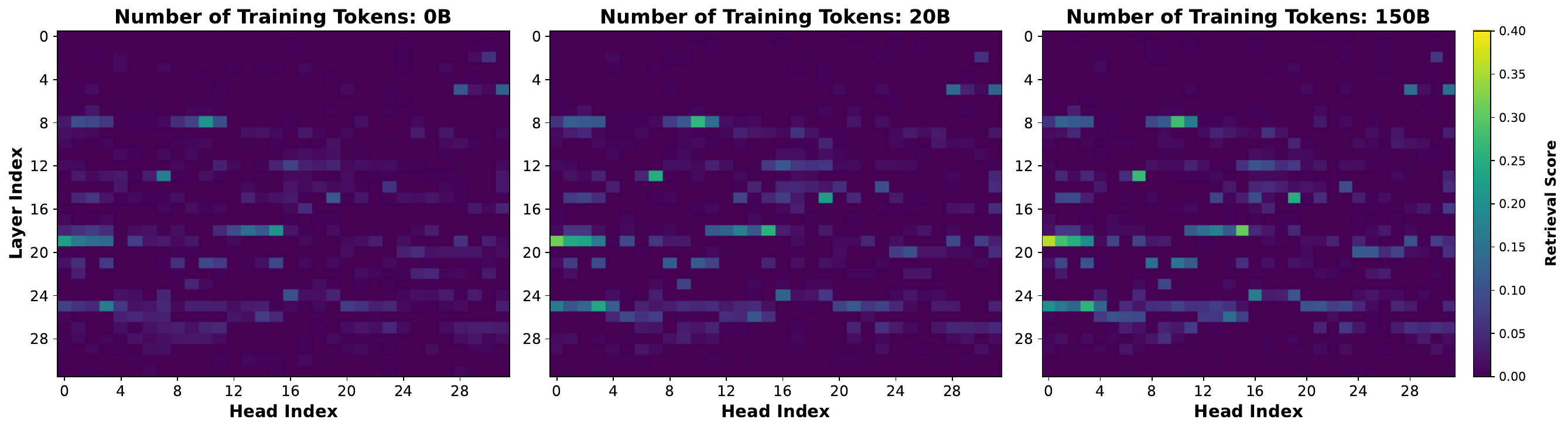}
    \caption{The retrieval scores of all attention heads across all layers of the Hunyuan-A13B base model at different stages of LCCP.}
    \label{figure: retrieval_score_map}
     
\end{figure*}

\begin{figure}[t]
    \centering
    \includegraphics[width=0.9\columnwidth]{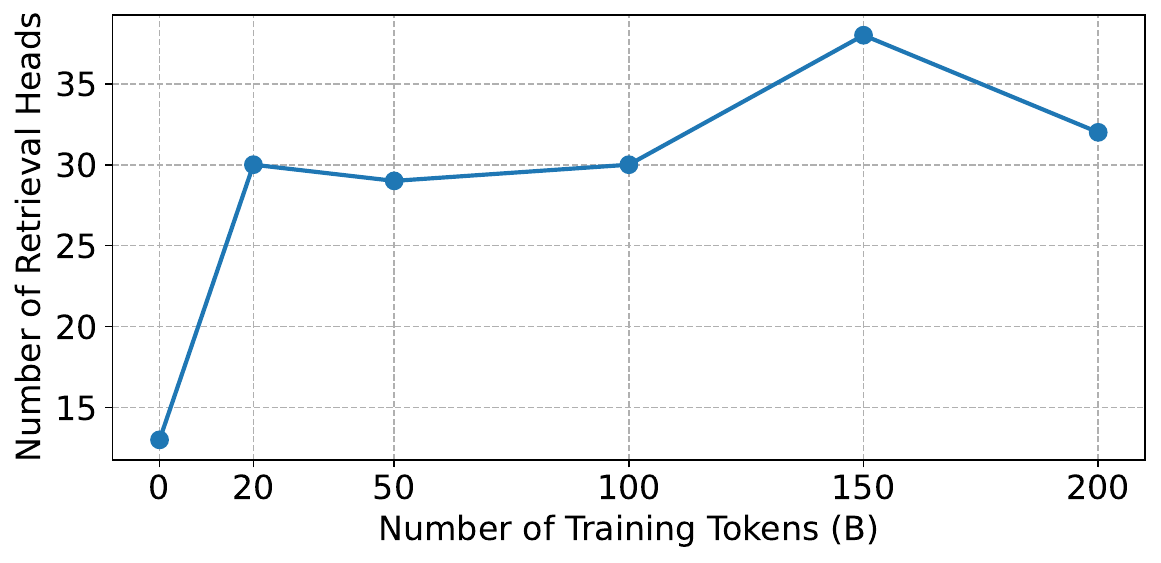}
    \includegraphics[width=0.9\columnwidth]{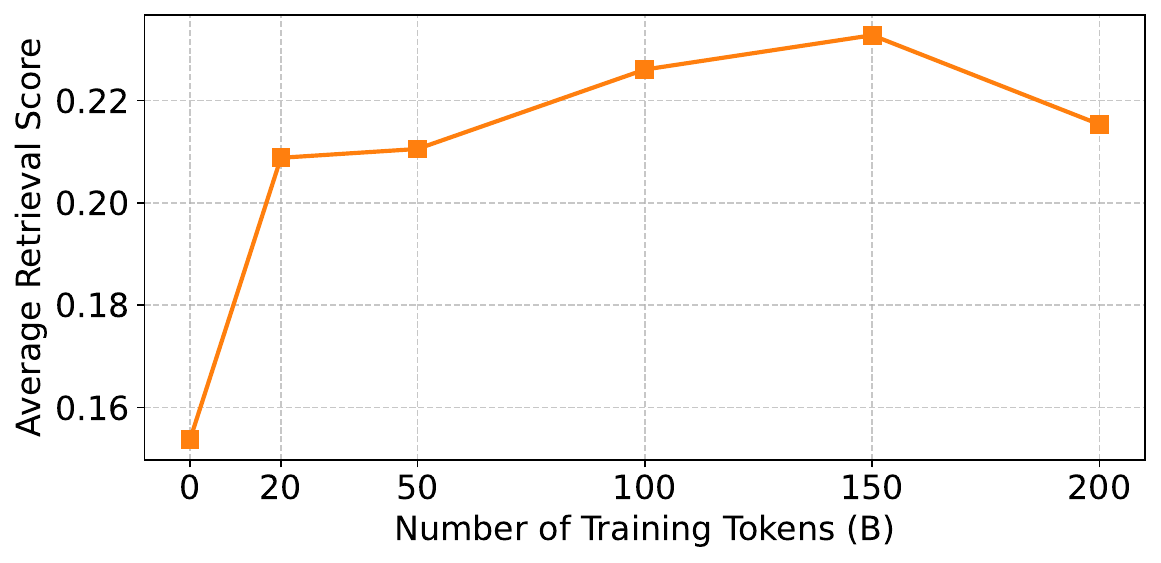}
    \caption{The number of retrieval heads and the average retrieval score of the initial retrieval heads of the Hunyuan-A13B base model at different stages of LCCP.}
    \label{figure: num_retrieval_heads}
\end{figure}

In this section, we dive into the attention mechanism to investigate how LCCP enhances the model's retrieval capability.
We follow the retrieval head identification method proposed by \citet{retrieval2025}.
By adapting the LongBio benchmark, we compute the retrieval scores of all attention heads in the Hunyuan-A13B base model at different training stages, and identify the specific retrieval heads responsible for retrieval and copy-paste behaviors for further analysis.
More details about retrieval score and retrieval head can be seen in Appendix~\ref{appendix: retrieval}.

\subsection{Evolution of Retrieval Heads}

We randomly sample instances from the LongBio single-needle test set and reformulate them into the NIAH input format described in Section~\ref{section: niah}.
Using the procedure described above, we compute the retrieval score of each attention head for each sample.
The final retrieval score for each attention head is obtained by averaging its retrieval scores across all samples.

We visualize the retrieval scores of all attention heads across all layers of the Hunyuan-A13B base model at different stages of LCCP using heatmaps in Figure~\ref{figure: retrieval_score_map}.
We also plot the curve showing the variation in the number of retrieval heads and average retrieval score of the initial retrieval heads with respect to the number of training tokens in Figure~\ref{figure: num_retrieval_heads}.

From the figure, it can be observed that as the number of training tokens increases, the number of retrieval heads also rises, and the average retrieval score of the model improves correspondingly.
Combined with our previous analyses on SFT, PPL, and NIAH, we conclude that LCCP enhances the retrieval capabilities and copy-paste behaviors of attention heads, specifically reflected in the increased retrieval scores of individual heads. This leads to a greater number of retrieval heads, thereby improving the model's overall retrieval and localization abilities, which ultimately impacts its long-context understanding after SFT.

\begin{table}[t]
\centering
\caption{The Pearson correlation coefficients between retrieval head metrics and performance on downstream SFT probe benchmarks.}
\label{tab:retrieval_correlation}
\small
\begin{tabular}{lcc}
\toprule
\textbf{Metric} & \makecell[c]{\textbf{\# of Retrieval} \\ \textbf{Heads}} & \makecell[c]{\textbf{Avg. Retrieval} \\ \textbf{Score}} \\ \midrule
RULER           & 0.8243                         & 0.8956                           \\
MRCR            & 0.8326                         & 0.8469                           \\
LongBio         & 0.5715                         & 0.6210                           \\ \midrule
\textbf{Average} & \textbf{0.7428}                & \textbf{0.7878}                  \\ \bottomrule
\end{tabular}
\end{table}

Besides, we evaluate the correlation between the retrieval head metrics, specifically the number of retrieval heads and their average scores, and the performance of the post-SFT model across various benchmarks, as summarized in Table~\ref{tab:retrieval_correlation}.\footnote{The significance testing can be seen in Appendix~\ref{appendix: p-value} Table~\ref{tab:retrieval_correlation_p_value}.}
Our results demonstrate a robust positive correlation between these internal retrieval head metrics and downstream task performance. Since these retrieval head metrics can be applied directly to the base model, \textbf{this finding establishes a computationally efficient method for monitoring the progression of LCCP, alleviating the need for resource-intensive SFT process and downstream evaluations}.

\section{Further Investigations \& Discussion}

In addition to our primary results, further investigation revealed several noteworthy observations.

\subsection{PPL Scaling Dynamics}

\begin{figure}[t]
    \centering
    \includegraphics[width=\columnwidth]{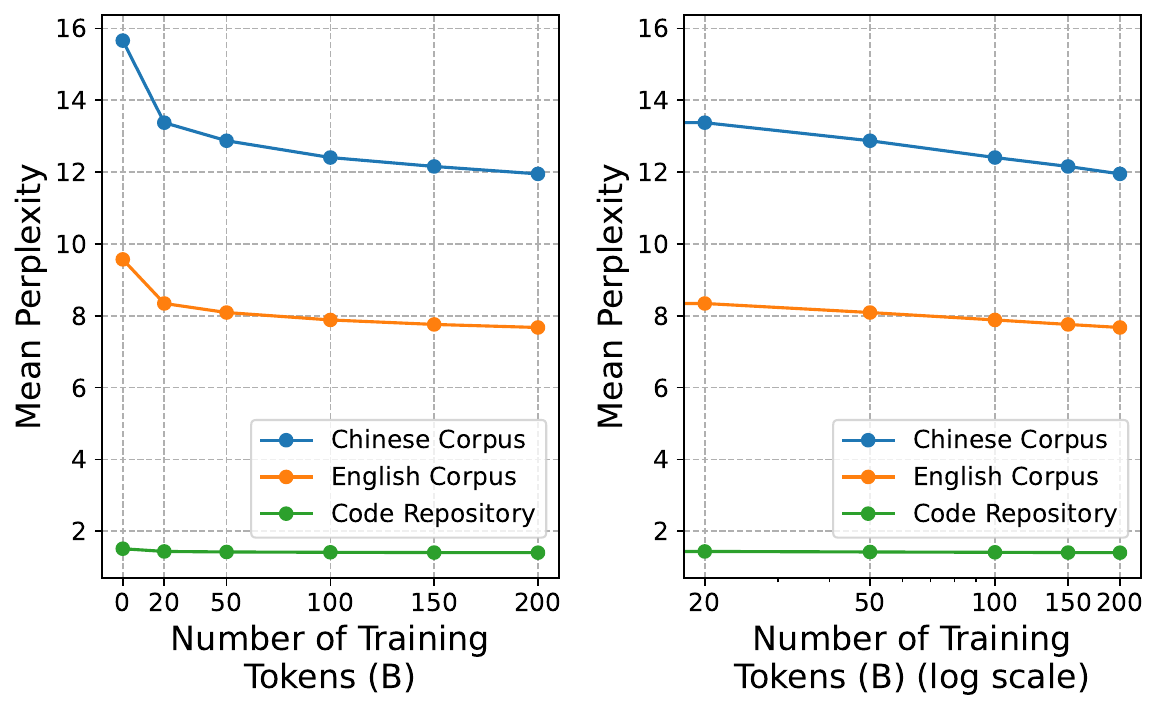}
    \caption{The PPL curve as a function of the number of training tokens of Hunyuan-A13B base model. The left panel displays the standard scale, while the right panel presents the logarithmic scale.}
    \label{figure: internal_corpus_ppl}
\end{figure}

To complement our probabilistic analysis, we also conducted a conventional PPL evaluation.
We prompt Hunyuan-A13B base models with carefully curated internal Chinese corpus, internal English corpus, and code repository, each consisting of 100 samples with a length of 64K tokens, and compute the corresponding metrics.
We ensure that all samples used for evaluation do not appear in the training data.

As shown in Figure~\ref{figure: internal_corpus_ppl}, with the increase in the number of tokens used for LCCP, the mean PPL across different corpora gradually decreases, and \textbf{the downward trend is consistent with the scaling law}.
PPL exhibits an approximately linear decrease with respect to the logarithm of the number of training tokens.
Based on these observations, we derive the following empirical formula:
\begin{equation*}
    PPL = A \cdot \log(N) + B
\end{equation*}
where $N$ denotes the number of training tokens in the LCCP phase, and $A$ and $B$ are parameters associated with the model architecture and the specific corpus used for PPL evaluation.
This behavior is largely consistent with the scaling law \citep{kaplan2020scaling, caballerobroken} and also aligns with the results of the earlier SFT analysis, indicating that as the number of training tokens increases, the gains in the model's long-context capability exhibit diminishing marginal improvements and gradually saturate.

\subsection{Mitigation of "Lost-in-the-middle"}

\begin{figure}[t]
    \centering
    \includegraphics[width=0.9\columnwidth]{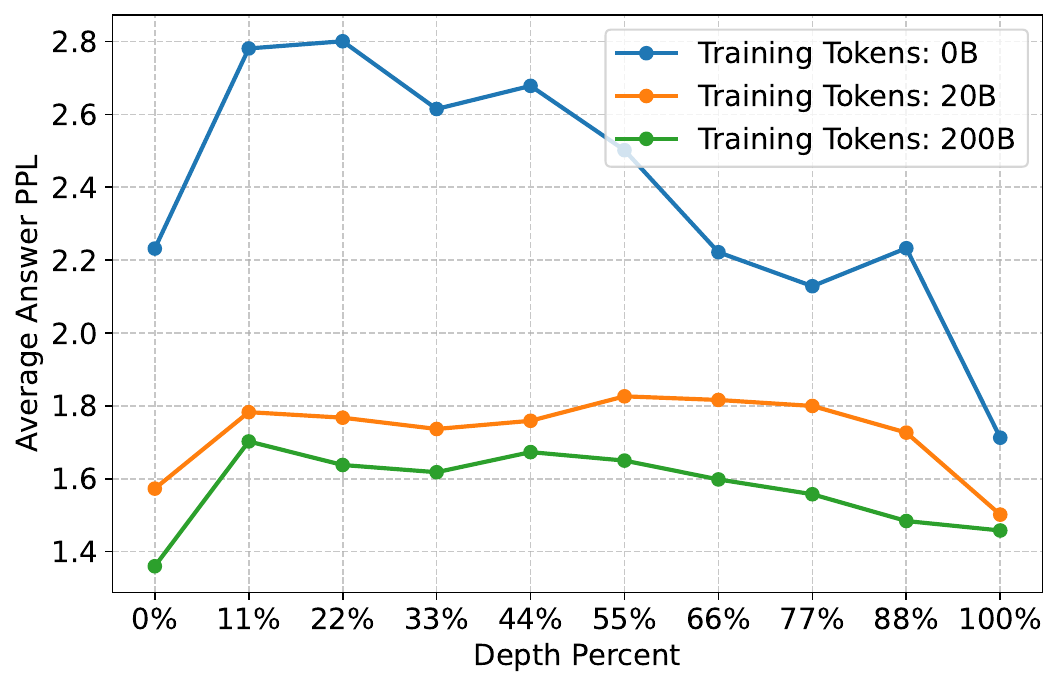}
    \caption{Average PPL across different document depths for Hunyuan-A13B base model at different stages of LCCP.}
    \label{figure: ppl_depth_curve}
\end{figure}

To conduct a more fine-grained quantitative analysis of Figure~\ref{figure: ppl_map}, we calculate the average PPL across different document depths, resulting in Figure~\ref{figure: ppl_depth_curve}.
As can be observed from the figure, \textbf{LCCP substantially enhances the model's retrieval capability and alleviates the "lost in the middle" phenomenon} \citep{liu2024lost}.
For sequences with lengths between 16K and 64K, the answer PPL is particularly high when the needle text is located at depths of 30\%-80\%, suggesting weaker retrieval performance for information embedded in the middle of long inputs (i.e., the "lost in the middle" phenomenon).
As the number of training tokens increases, the PPL of the answer tokens for input sequences longer than 40K tokens gradually decreases, indicating a progressive improvement in the model's retrieval capability.
In addition, the PPL for samples in which the needle text is located in the middle of the context is also substantially reduced, suggesting that the "lost in the middle" phenomenon is effectively alleviated.

\begin{figure*}[t]
    \centering
    \includegraphics[width=2\columnwidth]{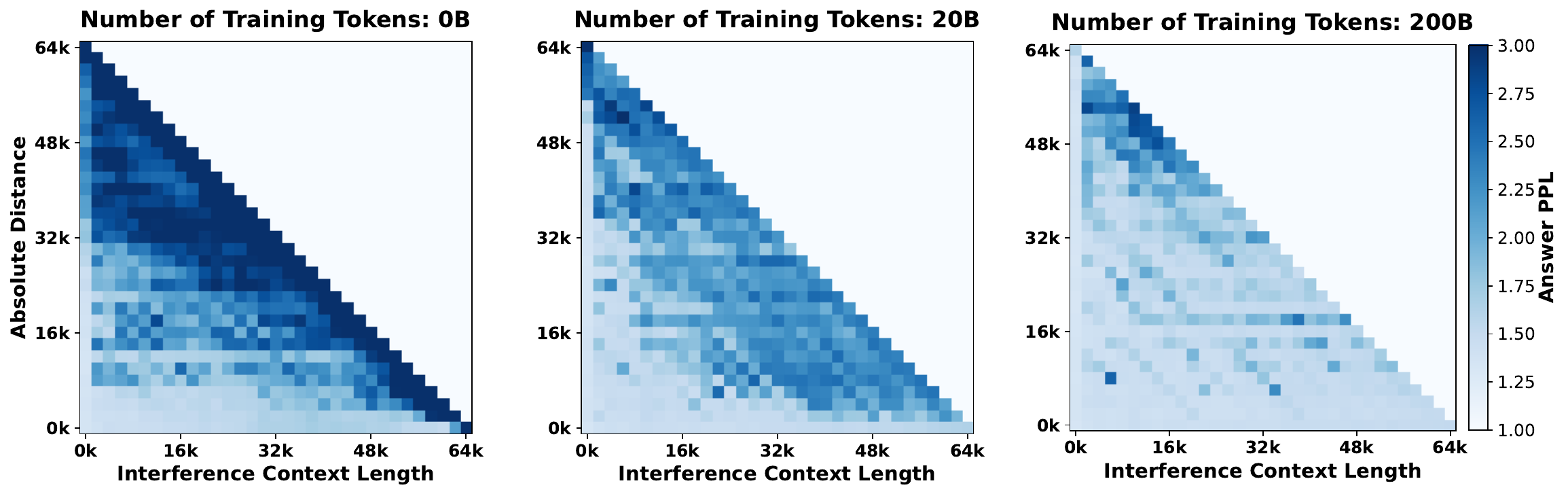}
    \caption{The PPL distribution for modified NIAH tasks across different context lengths and depths for Hunyuan-A13B base model at different stages of long-context continual pre-training. Each cell represents the average PPL of the answer tokens. Absolute Distance is the length of \textbf{\texttt{[Context After Needle Text]}} and Interference Context Length is the length of \textbf{\texttt{[Context Before Needle Text]}}.}
    \label{figure: abs_ppl_map}
\end{figure*}

\subsection{Robustness to Long-Context Interference}

\begin{figure}[t]
    \centering
    \includegraphics[width=0.9\columnwidth]{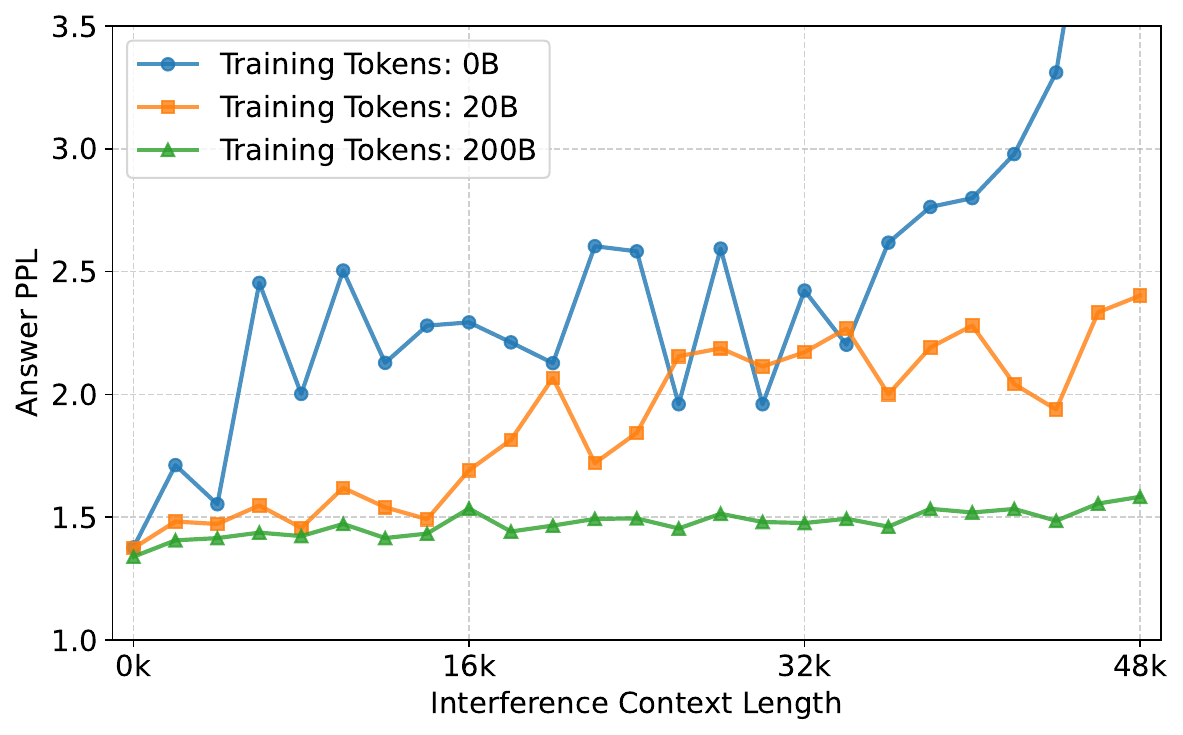}
    \caption{Answer PPL under increasing interference context length at a fixed absolute distance of 16K.}
    \label{figure: abs_ppl_curve_interference_16k}
     
\end{figure}

To investigate the relationship between the model's retrieval capability and the amount of distracting context in the input sequence, we design additional experiments.
We modify the original NIAH test samples by fixing the length of the \textbf{\texttt{[Context After Needle Text]}} (the absolute distance) while varying the length of the \textbf{\texttt{[Context Before Needle Text]}} (the interference context length), thereby systematically controlling the degree of contextual interference.
The results are shown in Figure~\ref{figure: abs_ppl_map} and Figure~\ref{figure: abs_ppl_curve_interference_16k}.

It can be observed that for models without LCCP, the PPL of the answer tokens increases as the interference context length grows (i.e., as contextual interference becomes stronger), indicating increased uncertainty and weak robustness to distraction.
In contrast, as the number of training tokens increases, the PPL variation across most absolute distance ranges becomes much less pronounced, and the PPL values remain consistently low under diverse test conditions.
This demonstrates that \textbf{LCCP significantly enhances the model's robustness to interference from long-context inputs}.

\subsection{Emergence of Retrieval Heads}

\begin{table}[t]
\centering
\caption{Overlap of top 30 retrieval heads and Spearman correlation between consecutive training stages.}
\label{tab:spearman_comparison}
\small
\begin{tabular}{ccc}
\toprule
\textbf{Training Stage} & \textbf{Overlap Heads} & \textbf{Spearman} \\ \midrule
0B $\to$ 20B            & 24/30                  & 0.7316            \\
20B $\to$ 50B           & 29/30                  & 0.9238            \\
50B $\to$ 100B          & 28/30                  & 0.9164            \\
100B $\to$ 150B         & 28/30                  & 0.8808            \\
150B $\to$ 200B         & 28/30                  & 0.8921            \\ \bottomrule
\end{tabular}
\end{table}

During our experiments, we observe an interesting phenomenon from Figure~\ref{figure: retrieval_score_map}: \textbf{the identity of retrieval heads is largely established during the initial pre-training phase, while subsequent LCCP primarily serves to amplify their inherent retrieval capabilities}.
To quantitatively investigate this phenomenon, we evaluate the overlap and consistency of retrieval heads across consecutive training stages. Specifically, Table~\ref{tab:spearman_comparison} reports the number of intersection heads between the top 30 heads (ranked by retrieval score) for adjacent checkpoints. Furthermore, we calculate the Spearman coefficient to measure the rank correlation of attention heads between these models, ordered by their retrieval scores.

Table \ref{tab:spearman_comparison} reveals that a minor shift in retrieval heads occurs during the initial 20B tokens of training (24/30). In subsequent stages, the overlap of the top 30 retrieval heads consistently remains above 93\%, indicating a highly stable functional identity. This high degree of overlap, coupled with strong Spearman correlations, suggests that the model's retrieval mechanism is largely determined at the pre-training stage. Consequently, the later phases of LCCP primarily serve to refine and amplify these established capabilities rather than reconfiguring the underlying functional heads.

\section{Conclusion}

In this paper, we presented a systematic investigation of the learning dynamics of LCCP using the industrial-grade Hunyuan-A13B model. By tracking a 200B-token training trajectory through a hierarchical framework encompassing behavioral, probabilistic, and mechanistic levels, we demonstrated that industrial-grade models require significantly more data, exceeding 150B tokens, to reach true saturation. Our analysis identifies the phenomenon of "deceptive saturation" in traditional benchmarks like NIAH and proposes PPL-based metrics and retrieval head monitoring as more reliable indicators of intrinsic convergence and training stability. These findings bridge the gap between small-scale experimentation and industrial application, providing both a robust evaluation methodology and a mechanistic roadmap for the predictable development and evaluation of long-context capabilities in industrial-grade LLMs.

\section{Limitation}

Due to the immense computational resources and data required for LCCP, we currently track the training trajectory of only 200B tokens during the LCCP process within the 64K context window. As more computational budget becomes available, we plan to further investigate the following dimensions:
(1) \textbf{Ultra-Long Context Scaling:} We aim to extend the context window beyond 256K tokens and scale the training data to over 200B tokens to verify if the saturation threshold and retrieval head stability persist at extreme scales.
(2) \textbf{Comprehensive Alignment Pipeline:} We will implement a full-scale post-training regime, including extensive SFT and RL, to examine how LCCP affects LLM's complex reasoning and instruction-following abilities.
(3) \textbf{Systematic Ablation of Recipes}: We plan to conduct rigorous ablation studies on data mixture ratios and hyper-parameters, such as RoPE base frequencies, to quantify their precise impact on long-range dependency.
(4) \textbf{Mechanistic Intervention and Optimization:} We intend to explore whether the identified retrieval heads can be actively intervened upon or optimized during training to further improve retrieval efficiency and model factuality.
(5) \textbf{Cross-Model Generalization:} We plan to extend our LCCP investigation to other open-source LLMs, such as the DeepSeek and Qwen series, to validate the universality of our hierarchical monitoring framework and learning dynamics across diverse architectures.

\bibliography{custom}

\appendix

\section{Appendix}
\label{sec:appendix}

\subsection{Retrieval Score \& Retrieval Head}
\label{appendix: retrieval}

\citet{retrieval2025} introduces a retrieval score to measure the frequency of an attention head's copy-paste behavior during autoregressive decoding.
An attention head with a high retrieval score suggests that statistically across various contexts, this head is frequently copying the input tokens from the input to the output.

For each input $x$ formatted as the NIAH input described in Section~\ref{section: niah}, we feed it into the model for forward computation and obtain the attention scores of each attention head $a \in \mathbb{R}^{|x|}$.
An attention head $h$ copies and pastes a token from the needle text $k$ to the answer text if it follows two criteria: (1) $w \in k$, i.e., $w$ is a token within the needle sentence. (2) $x_j = w, j = \arg \max(a), j \in i_q$, i.e., the input token that receives the most attention probability mass by this head is a token within the needle and is the same token as the currently generated token.
Let $g_h$ be the set containing all tokens copy and pasted by a given head $h$, the calculation for the retrieval score for head $h$ for a single sample can be formulated as:
\begin{equation}
    \text{Retrieval score for head } h = \frac{|g_h \cap k|}{|k|}
\end{equation}
Intuitively, the retrieval score represents a token-level recall rate of the most attended tokens by an attention head.
If the retrieval score of an attention head exceeds a threshold $r$, we consider it to be a retrieval head.
In our experiments, we set the threshold $r$ to 0.1.

\subsection{Significance Testing}
\label{appendix: p-value}

Table~\ref{tab:naih_ppl_correlation_p_value} shows the  p-value of Pearson correlation coefficients between NIAH metrics and performance on downstream SFT probe benchmarks.
These results indicate that the NIAH PPL metric correlates significantly ($p < 0.05$) with SFT performance on RULER and MRCR, while the correlations with LongBio, though not statistically significant, still trend positively. PPL-based metrics can reflect downstream capabilities when the evaluation corpus and computation method are carefully designed, precisely the rationale behind introducing NIAH PPL. More importantly, analyzing the LCCP process requires a holistic view that combines multiple metrics and SFT evaluations, rather than relying on any single indicator.

\begin{table}[t]
\centering
\caption{The p-value of Pearson correlation coefficients between NIAH metrics and performance on downstream SFT probe benchmarks.}
\label{tab:naih_ppl_correlation_p_value}
\small
\begin{tabular}{lcc}
\toprule
\textbf{Metric} & \textbf{NIAH Score} & \textbf{NIAH PPL} \\ \midrule
RULER               & 0.0299              & 0.0114           \\
MRCR                & 0.0166              & 0.0087            \\
LongBio             & 0.3151              & 0.1816           \\ \bottomrule
\end{tabular}
\end{table}

Table~\ref{tab:retrieval_correlation_p_value} reports the p-vaue of Pearson correlation coefficients between retrieval head metrics and performance on downstream SFT probe benchmarks.
We observe significant correlations ($p < 0.05$) between the proposed retrieval head metrics and SFT performance on the RULER and MRCR benchmarks, with a similar positive trend observed for LongBio. This strong correspondence between internal metrics and downstream capabilities allows for efficient monitoring of LCCP. By applying these metrics directly to the base model, researchers can track pre-training progress without the computational burden of intermediate SFT or resource-intensive downstream assessments.

\begin{table}[t]
\centering
\caption{The p-vaue of Pearson correlation coefficients between retrieval head metrics and performance on downstream SFT probe benchmarks.}
\label{tab:retrieval_correlation_p_value}
\small
\begin{tabular}{lcc}
\toprule
\textbf{Metric} & \makecell[c]{\textbf{\# of Retrieval} \\ \textbf{Heads}} & \makecell[c]{\textbf{Avg. Retrieval} \\ \textbf{Score}} \\ \midrule
RULER           & 0.0436                          & 0.0158                               \\
MRCR            & 0.0397                          & 0.0334                               \\
LongBio         & 0.2361                          & 0.1882                               \\ \bottomrule
\end{tabular}
\end{table}

\end{document}